# Natural Language Processing, Sentiment Analysis and Clinical Analytics

**Adil Rajput**

**(aiilahi@effatuniversity.edu.sa)**

*Assistant Professor, Information System Department, Effat University*

*An Nazlah Al Yamaniyyah, Jeddah 22332, Jeddah, Saudi Arabia*

## Abstract

Recent advances in Big Data has prompted health care practitioners to utilize the data available on social media to discern sentiment and emotions' expression. Health Informatics and Clinical Analytics depend heavily on information gathered from diverse sources. Traditionally, a healthcare practitioner will ask a patient to fill out a questionnaire that will form the basis of diagnosing the medical condition. However, medical practitioners have access to many sources of data including the patients' writings on various media. Natural Language Processing (NLP) allows researchers to gather such data and analyze it to glean the underlying meaning of such writings. The field of sentiment analysis – applied to many other domains – depend heavily on techniques utilized by NLP. This work will look into various prevalent theories underlying the NLP field and how they can be leveraged to gather users' sentiments on social media. Such sentiments can be culled over a period of time thus minimizing the errors introduced by data input and other stressors. Furthermore, we look at some applications of sentiment analysis and application of NLP to mental health. The reader will also learn about the NLTK toolkit that implements various NLP theories and how they can make the data scavenging process a lot easier.

## 1. Introduction

The Big Data revolution has changed the way scientists approach problems in almost every (if not all) area of research. The Big Data field culls concepts from various fields of Computer Science. These include Natural Language Processing, Information Retrieval, Artificial Intelligence, Machine Learning, network analysis and graph theory to name a few. The aforementioned fields have been part and parcel of research in Computer Science for many decades. However, the advent of Web 2.0 and social media resulted in the three Vs of the big data - Variety, Veracity and Volume (Laney et. al., 2001).

### 1.1. NLP and Healthcare/Clinical Analytics

One of the challenges that researchers and practitioners face in both psychology and psychiatry is access to data that truly reflects the mental state of the subject/patient. The traditional approaches depend on gathering data from the subject and the immediate family/friends and/or asking select individuals belonging to a certain group to fill put surveys/questionnaires that might provide an insight into mental state of various individuals/groups.

Sentiment analysis domain – also known as Opinion Mining – allows scientists to sift through the text gathered via various sources and glean how the subject at hand feels. The area depends heavily on techniques in Natural Language Processing (NLP). The Natural Language Processing allows a machine to process a natural human language and translates it to a format that the machine understands. NLP dates back to the 1960s but became very popular with the advent of the World Wide Web and search engines. The query processing capabilities of search engines required to add context to the terms entered by users and in turn present a set of results that the user can choose from.

## 1.2. Sentiment Analysis

Utilizing the techniques from NLP, sentiment analysis field looks at users' expressions and in turn associate emotions with what the user has provided. The cultural norms add a different twist to this area. For example, the following statement could be interpreted very differently.

<p style="text-align:center">"This new gadget is bad!"</p>

While the obvious meaning alludes to the user's dislike of the gadget, user community belonging to a certain age group would consider the above statement as a resounding endorsement of the gadget at hand. Furthermore, the sentiment analysis looks at the time at which the user expressed the sentiment or opinion. The same user can be under certain stressors which can cloud their judgment and hence gathering statements on a time continuum can provide better assurance of the sentiments expressed.

The social media platform provides both challenges and opportunities in this area. On a positive note, it grants anonymity to the person writing on the web thus allowing him/her to express their feelings freely (Rajadesingan et. Al., 2015). Moreover, the data can be gathered for a given interval that can prove vital in ensuring consistency. The data gleaned in this manner will offer a preponderance of evidence supporting the researcher's hypothesis and can provide a solid foundation for scientific deductions. Gathering data from the web has become the choice of many fields such as Marketing etc. Google, YouTube and Amazon are examples of how companies can provide customized content to the end-user. Many such fields can depend greatly on objective metrics such as number of likes, total number of items sold given an age range etc. However, the fields of psychology/psychiatry do not have such luxury as the data is in form of text written by users on the various media such as blogs, social media etc. This dimension adds more complexity due to a) use of different languages on a certain topic/blog, b) use of non-standard words that cannot be found in a dictionary and c) use of emojis and symbols. These questions are tackled by experts in the Natural Language Processing (NLP) domain along with those working in the sentiment analysis area.

There is a need of providing social scientists and psychiatrists the requisite vocabulary and the basic tools to scavenge data from the web, parse it appropriately and glean the contextual information. This work is intended to be a step in this direction. Specifically, the paper provides the following:

1. Provide a basic understanding on various prevalent theories in Natural Language Processing (NLP)
2. Explain the traditional and the statistical approaches to NLP
3. Look at the work done in the area of sentiment analysis and the challenges faced in the light of mental health issues

4. Present a brief synopsis of various applications in applying NLP concepts to mental health issues and sentiment analysis

The paper will introduce the key concepts and definitions pertaining to each section rather than lumping it altogether in a separate section.

## 2. Natural Language Processing

The field of NLP dates back to few decades and has matured quite significantly over the years. Initially confined to gathering data from a limited set of digitized documents, the advent of World Wide Web saw an explosion in information in many different languages. Significant amount of work was done in the information Retrieval (IR) field which is considered an application of the Natural Language Processing domain. Before discussing the IR techniques, a bit more, let us delve into the theoretical and practical aspects of NLP.

### 2.1. Traditional Approach - Key Concepts

Initially, the NLP approach followed the following discrete steps.

1. Text Preprocessing/Tokenization
2. Lexical Analysis
3. Syntactical Analysis
4. Semantic Analysis

### 2.1.1. Preprocessing/Tokenization

The first challenge is to segment a given document into words and sentences. The word token - initially confined to programming languages theory - is now synonymous with segmenting the text into words. Most of the languages use the white space as the delimiter but it can be a bit tricky in certain languages. While seeming straightforward, the challenges include separating words such as 'I'm' into 'I am' and deciding whether or not to separate a token such as 'high-impact' into two words. Complicating the matter further would be the language of the document. The unicode standard helps tremendously as each character is assigned a unique value and therefore makes it practical to decide upon the underlying language.

Another concept that NLP experts use quite often is "Regular Expression (RE)". Also finding its root in the computer programming language theory, RE specifies the format of the string that needs to be looked at. As an example, a password string (token) that can contain upper case letters only will be specified as [A-Z] while a string counting numbers will be specified as [0-9]. The importance of RE will become apparent in the next subsection.

In addition to segmenting the text into token/words, the NLP domain places great emphasis on finding the boundary of sentences. While many languages will use punctuation marks to define sentence boundaries, other languages such as Chinese, Korean etc. prove to be much more difficult in this regards. Complicating the matter further are short forms that use the period symbol '.'. While used for ending a sentence, a token such as 'Mr.' might send the wrong signal.

## 2.1.2. Lexical Analysis

After processing the text, the next challenge is to divide the text into lexemes. A lexeme in linguistics represents a meaning and is considered the unit of lexicon. A lexeme can have different endings - known as inflectional endings. As an example, the term 'sleep' is the unit that can take various forms such as 'sleeping', 'slept' or 'sleeps'. The unit token is also known as lemma. A lexeme is composed of morphemes - bound and unbound/free. An unbound morpheme are 'tokens' that can be independent words such as cat. Bound morphemes are affixes and suffixes such as 'un', '-tion' etc.

After preprocessing of text and segmenting it into words, the NLP practitioners would take each token and reduce it to its unit lexeme form. Thus, the words 'depression' and 'depressed' will both be reduced to one-unit form - 'depress'. This process is also known as stemming where each token is reduced to a root form called stem. This term is more prevalent in Computer Science and in certain cases might not be the same as the lemma. The most famous algorithm for this technique is the Porter algorithm (Porter et. al., 1980) which was later improved to Porter2 or Snowball algorithm. Lancaster Stemmer is also used frequently but is considered a bit more aggressive. This will be discussed in more detailed in the practical section.

One of the most important benefit of stemming is to gather a frequency distribution of various words in a given text. The frequency distribution helps surmise the topic of the text being considered at hand. The famous tf-idf algorithm in Linguistics and Computer Science is widely used. The tf measures the frequency of the terms present in the document and infers the subject/keywords describing the document. The idf factor focuses on eliminating the commonly used words such as prepositions, articles etc. allowing the tf factor to accurately represent the subject of the document at hand. While many other techniques have been proposed and tested, tf-idf algorithm is usually the starting point when dealing with texts.

## 2.1.3. Syntactical Analysis

Now that we understand how a text can be broken down into sentences and words using the concept of tokens, the next challenge is to ensure that the text being processed is following rules of grammar and is conveying certain meaning. Syntactical analysis is the process that ensures that rules of grammar are being followed. As an example, consider the sentence "Mary Joe road deer drive.". The tokens and the period will indicate a full sentence but does not convey any meaning. The grammars are described as sets of rules. The following rules for example, describe the rules for representing numbers and the four operators namely addition, subtraction, division and multiplication.

<E> -> Number

<E> -> (<E>)

<E> -> <E> + <E>

<E> -> <E> - <E>

<E> -> <E> / <E>

<E> -> <E> * <E>

The Grammar (referred to as mathematical grammar) is composed of terminal and non-terminal symbols. In the above example, Number is a terminal symbol while <E> is a non-terminal symbol. If we assume that the Number symbol represents integers, then the following expressions when parsed will conform to the above grammar.

$$134 + 256, 134, (256)$$

However, expressions such as '-134', '134' '25' and '134 / 12 34' will not conform to the grammar described above. The process of ensuring the tokens follow a particular grammar is also referred to parsing by Computer Scientists / Computational Linguists. Both the lexical analyzer described in the previous subsection and a parser is needed to process text. While the above might seem overwhelming a bit at first, think of it as this: If the text has the following two sentences, how can it be decided that the sentences are conforming to English grammar?

"The dog ran after the ball"

"The ball dog ran ball the"

A human looking at the two sentences above will dismiss the second sentence as gibberish right away, but it is not so easy for a computer to discern. The question computer scientists and computational linguists traditionally faced was whether or not a natural language can be represented by a mathematical grammar (also referred to as formal grammar). The grammars that have been traditionally the subject of researchers are best described by Chomsky hierarchy (Chomsky et. al., 2012). The formal grammars can be divided as follows:

1. Unrestricted grammars: These are grammars that would have a rule like α -> β where α and β can both be terminals, non-terminals or null. Such The unrestricted grammars are the most general and include all the remaining grammars. The problem with such grammars is that they are too general to describe any programming or natural language.
2. Context-Sensitive grammars: These grammars are described by rules such as αAβ -> αγβ where α, β can be non-terminals, terminals or empty, γ can be can be non-terminals or terminals but never empty, and A has to be a non-terminal. In simple terms, the context-sensitive grammars refer to the fact that certain words can only be appropriate in a certain context - a problem that is intuitive to humans. The issue with such grammars are that they are extremely difficult computationally (if decidable at all). Note that Context-Sensitive grammars contain the Context-Free grammars and Regular grammars but not vice-versa.
3. Context-Free grammars: These grammars are described by the rules such as A -> γ where γ can be can be non-terminals or terminals but never empty, and A has to be a non-terminal. These grammars are used to describe the syntax of most programming languages such as C etc. The Context-Free grammars contain the regular grammar but not vice versa.
4. Regular grammars: These grammars are described by A -> aB or A -> Ba where a is a terminal and both A and B are non-terminals. The regular grammars are used to define the search patterns and lexical structure of the programming languages.

The problem researchers in Computer Science and Computational Linguistics faced for the longest time was that while the above was enough to describe the programming languages, it was not sufficient for natural languages. This was addressed by Statistical approach as we shall see in section 2.2.

## *2.1.4. Semantic Analysis*

Finally, we will briefly discuss the semantic analysis before taking a look at the statistical. Recall that when we want to process a text, we need to preprocess the text where we break down the text into words and sentences. Next, we perform a lexical analysis where we will group various words who have the same root token called lemma together. The syntactical parsing allows us to ensure that the text is following a grammatical structure and hence can be part of a given language. Also recall that most of the languages can be processed by this approach while few languages such as Chinese, Thai etc. face difficulty in the process of tokenization and lemmatization. Once this is accomplished, we need to ensure whether or not the sentence written is conveying a meaning. In addition to the example in Syntactical analysis where one sentence was termed as gibberish, consider the following sentences:

"I am going down"

"I am feeling down"

"I am walking down"

The three sentences can be interpreted differently. Moreover, the first and third sentence could mean that the person is going to a floor down or the first sentence could mean the person is about to have a flu if the symptoms of flu were discussed prior to this sentence (recall the context-sensitive grammar where the text requires history of the text). It can also mean a player mentioning that he/she might lose the game. In summary, one can surmise that there are many possibilities that any given sentence can convey. Linguistics over the past century has seen many theories crop up that have been the basis for the work of Computer Scientists/Computational Linguists. While covering all the theories here is beyond the scope of this work, we will briefly summarize four such theories here.

1. Formal Semantics: The key premise in formal semantics is that there is no difference between natural and artificial languages. Both can be represented as a set of rules and based on such rules we can make deductions. As an example, consider the following rules:

"Every man is mortal"

"John is a man"

"John is Mortal"

This can be represented mathematically as follows:

Man -> Mortal

Man(John)

==> Mortal(John)

2. Cognitive Semantics: As opposed to the formal semantics, cognitive semanticists believe in intuition/psychological aspect of the communication. In other words, they argue that each sentence has an intuitive aspect that delivers the message. For example, "He is going down" can be interpreted as becoming sick or the physical action of heading down. The difference lies in the context/intuition. Furthermore, various cultures can add meanings to various sentences
3. Lexical Semantics: The lexical semantics deal with the meanings of individual lexemes and the meanings entailed by them. The lexemes can have suffixes and affixes and they can alter the meaning of the individual word. Moreover, the individual lexemes might have sensory meaning associated with them. For example, the following sentences are correct in terms of grammar but the second one will not be deemed a correct sentence.

    "The cat chased a mouse"

    "The mouse chased a cat"

4. Compositional Semantics: The compositional semantics do not look at the individual meaning of the lexemes but rather look at how a sentence is composed. For example, a sentence can be composed of a noun phrase or a verb phrase. So the following two sentences will be considered correct:

    "Jack is a boy"

    "J is a B"

The key premise behind the above is that minus the lexical parts what remains are the rules of composition.

The traditional approach to processing text yields descent results when it comes to preprocessing and tokenization phase. However, one can surmise from the examples above that the task increases in complexity in the syntactical and semantical analysis phase. This has given rise to the statistical approach to NLP which will be discussed later in this paper. However, understanding the above concepts are paramount to comprehending and implementing the statistical approaches.

## 2.2. Statistical Approach - Key Concepts

As we saw in the previous section, the traditional approach has its share of challenges when performing the syntactical and semantical analysis. The statistical approach takes its motivation from the machine learning (ML) approach. Simply put the ML approach takes a subset of data and studies the underlying structure and behavior of the input and the output. Specifically, the process finds the optimal way to convert the given input to the desired output - known as 'supervised learning'. The data utilized in the supervised learning is known as the training dataset. Once the algorithm is discovered, the algorithm is applied to a new dataset - test dataset - to see the effectiveness of the algorithm. This process is termed as 'unsupervised learning'. While many complexities underlie the above process, the following concepts describe the key ideas in this approach.

## 2.2.1. Corpus and its intricacies

While many definitions exist for corpus, we chose the following from (Sinclair, 1991):

> "A collection of naturally occurring text, chosen to characterize a state or variety of a language"

For NLP purposes, the text needs to be machine readable, so it can be annotated. The annotation process in NLP takes a text and adds special tags known as metadata to various words - described in more detail in a later section.

Researchers over the past decades have provided us with many corpora. These include the Brown corpus (Marcus et. al., 1993), British National corpus (Aston et. al., 1998), International corpus of English and Google Ngram corpus (Lin et. al., 2012). Such corpora relieve us from the legal aspect as pointed out in (Chang et.al., 2016). However, the choice of corpus is important given the task at hand and the results can be highly domain specific (Gordon et. al., 2009). Intuitively, someone working intending to study British population would naturally look at the British National corpus to get better insights. Despite having many copra available to us, there is a constant need to build new corpora. For example, (Rajput et. al., 2018) describes the need to have a corpus for psychology and psychiatry. This begs the question: What are the key characteristics of a corpus? The size, balance and representativeness are three aspects that need to be looked at.

### 2.2.1.1. *Size*

The very first question that needs to be answered is how big a corpus should be in order to represent the desired text. Since corpora depend on sampling, the answer to this question will help build a corpus that can provide the researchers they are looking for. While intuitively it might make sense to keep the corpus as large as possible, having a small corpus fulfills a very important purpose - performing annotation and studying grammatical/underlying text structure. Thus, someone focused on annotating a given text heavily and/or studying the grammatical structure of a particular text would find it very difficult if not impossible to work with a corpus that is too big. Loosely speaking a large corpus helps in studying the occurrences of lexemes, their frequencies and concordances of various tokens (Berber-Sardinha, T. et. al., 2000). A concordance is when certain words occur together in certain text. For example, when Google fills certain words such as 'have' when someone types 'I can' is based on the study of various queries and corpora. This will become more evident when we discuss Part-of-Speech tagging.

### 2.2.1.2. *Balance*

The Balance of a corpus refers to the ability of a corpus to represent the language being studied. Even before the era of acronyms of the chat era, one would expect that the text scripts of spoken language would be quite different from the written texts. Furthermore, languages having more than one spoken dialect will have different text scripts representing different regions. As an example, Arabic speaking community in Tunisia will have different choices of word/lexemes compared to someone from Egypt. Adding to the complexity is the choice of colloquial words specific to that particular community. Furthermore, a particular lexeme can have different connotations across different communities. As an example, consider the word 'unionized' that can be pronounced as "union-ized" (specific to unions) or 'un-ionized' (specific to Chemistry). Similarly, the acronym ROE will be read as 'Return on Equity' by the finance community while the military community would read it as 'Rules of Engagement'. As one can imagine, various corpora are specific to different domains and hence the continuous need to build corpora.

### *2.2.1.3.     Representativeness*

Consider studying/tagging a Shakespeare act to represent how a daily dialogue commences between people. Obviously, this will not be representative of the times we live in. A better example would be to revisit the way people chatted when instant messaging made its debut. With the advent of time, terms such as 'lol' (laugh out loud), 'imo' (in my opinion) etc. have been added to the everyday jargon. So how can a text continue to represent a language over a period of time? How often it needs to be updated? These are the questions one needs to keep in mind when building a corpus. However, there are corpora which should not be updated at all. As an example, the text scripts representing the phone records of conversation during the nineties represent the choice of words in that decade. On the other hand, the advent of social media and various terminologies/acronyms require a frequent update to certain corpora. For example, should a corpus representing behavior of various psychological disorder be limited to answers of individuals selected for a survey? or should it incorporate lexemes/words that are gleaned from various tweets discussing depression? How should one deal with the Out of Vocabulary (OOV) words that cannot be found in a dictionary but are prevalent among such communities? While answers to such questions are subject to scrutiny by many researchers and beyond the scope of this tutorial, they underlie the importance of corpus building and selection in human, social and medical sciences.

## *2.2.2. Part-of-Speech(POS) Tagging*

So far, we have discussed creating the corpus which will be either our training or test dataset or both. The Part-of-Speech tagging is a process akin to what kids are taught when studying any grammar namely recognizing the type of word they come across. For example, the sentence "The dog jumped over the fence" can be tagged as the following:

"The/Determiner (Definite Article) dog/Noun jumped/Verb over/Preposition the/Determiner fence/Noun"

Let us make few observations. First of all, while the above seems easy (once we learn the grammatical rules), the problem is much more complex in NLP as many words van have different possible tags. The word 'over' can be a preposition (as shown above) or an adverb (consider the sentence 'He fell over'). Secondly, we would recognize the word 'jumped' as a verb after the process of lemmatization as described earlier in this primer. As we noted earlier, the preprocessing and lexical analysis process specific to traditional approaches is equally beneficial and applicable to the statistical approach. Thirdly, in the corpus creation we discussed that all corpora are especially helpful in annotation and POS tagging. A small corpus usually would represent the rules of a given language quite comprehensively and hence the benefit of a small corpus (Brants, 2000). We will also discuss the application of taggers in the next section which will discuss annotations in some detail.

There are three main prevalent techniques for POS tagging widely used.

1. Rule-Base POS Tagging: The rule-based tagging where manual rules are encoded. For example, one rule could be that a noun always follows a determiner. This can be very helpful if consider the word 'excuse' which can be both tagged as a noun or a verb. Now consider the following two sentences:

"The excuse was not accepted"

"The employee was not excused"

A rule-based tagger would thus tag the 'excuse' as a noun. The problem with rule-based taggers is that it takes many iterations over a text and hence computational time and space. To alleviate this problem, researchers take a probabilistic approach know as Markov Model approach.

2. Markov Model POS Tagging: The Markov model is a concept in probability and statistics which states that events are not always independent (such as tossing a coin where each outcome is independent of the previous one). Rather, the events might be related to the history. Markov models in its simplest form assume that given a number of sequential events say $E_1$, $E_2$, …. $E_n$, the next event $E_{n+1}$ is simply dependent on the previous event. This is known as the memoryless property. As an example, some would argue that the probability of a basketball player making the second free throw goes up if he makes the first one. Conversely, the probability will go down if the player misses the first shot.

Revisiting the example from the rule-based tagging, the probability of a noun following a determiner can be set to 1. Hence the word 'excuse' will be tagged as a noun in the first sentence. This can be very helpful in ensuring that the sentence follows the grammatical rules of a particular language. Markov Model makes the assumption that the information about the last state is available. In certain cases, more history is available as is the case in speech tagging. Specifically, we can look for other words that appear in the sentence on a high frequency basis. Hidden Markov Models (HMM) is the mathematical model that interest's researchers and there are many variations that have been tried and tested such as Variable Markov Memory Models. Taking it a step further, researchers also look for words that frequently occur together in a document - a concept known as concordances.

3. Feature-Base POS Tagging: Certain features of a language can help the tagging process further. Such features are known apriori such as proper names are always capitalized. These features can become quite complex and have been the subject of researchers for quite some time. Features help in providing context to the tagging process. For example, the word 'Citibank' would provide the context that the document at hand is probably related to Finance.

The tagging approaches help NLP practitioners in gleaning information about the document at hand. The metadata can include the author name, the subject matter, date of publishing etc. Regardless of the choice of taggers employed (or combination of them), POS tagging offers a great value to NLP practitioners. One such application is 'Annotation'

## *2.2.3. Treebank Annotation*

Simply put, the annotation process takes a text corpus and attaches metadata information to the text. Many such corpora are available and have been developed over the years. The Brown

Corpus developed at Brown University is heavily annotated as are many other corpora. The annotation traditionally was done manually but many computational algorithms now exist to perform the annotation automatically. Recall from the Syntactical analysis section that a text must conform to a particular grammar. While the traditional approaches had limited results approaching this problem, POS tagging and treebank annotation improves the grammatical accuracy quite significantly.

The key premise behind treebank annotation is that it views a sentence as a tree. The tree can be built using a constituency-based annotation or dependency-based annotation (Marcus et. al., 1993). A constituency-based approach divides a sentence to a Noun Phrase (NP) or a Verb Phrase (VP) as shown below.

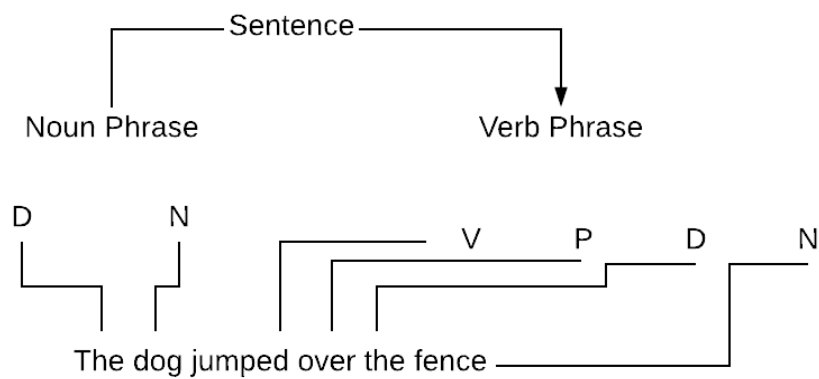

*Figure 1 Constituency Based Annotation*

A dependency-based annotation on the other hand focuses on the verb and builds the tree around it. Such an approach is very helpful for certain languages such as Arabic.

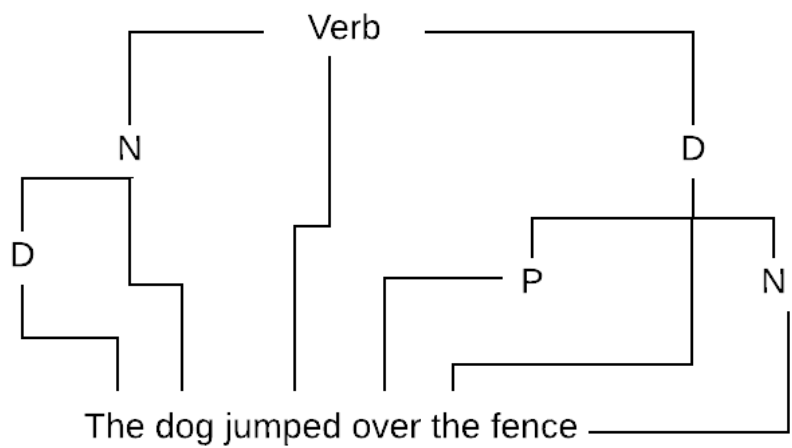

*Figure 2 Dependency Based Annotation*

Both the constituency-based and dependency-based annotations have pros and cons. While we will defer this discussion to other work in the future, treebank annotations address many problems in semantic analysis that we saw in the traditional approach.

## 3. Applications

Work done by (Landers et. al., 2016) provides a framework that helps in extracting Big Data from the Web. The authors specifically promote a "theory-driven" web scraping where researchers are encouraged to formulate questions and hypotheses before embarking on scavenging data from the web. This can include various criteria such as age, demographics etc. The authors put forward a case study in which the gender identity is deduced and prove a hypothesis regarding women behaviour. Lastly, the authors point out the importance of web crawlers work and how Application Programming Interfaces (APIs) plays in the Big Data field. Most if not all the social media platforms provide APIs that help the researchers gather data.

Both NLP and Sentiment Analysis have been applied in many areas such as marketing etc. However, both social sciences and medicine have just started to feel the impact. Researchers have recently started looking at how big data techniques can be applied to mental health issues such as detection of depression. In this section we present an overview of various applications both in the field of sentiment analysis and detecting mental health issues.

### 3.1. **Sentiment Analysis**

The basic application of sentiment analysis lies in gathering the opinion of people. Such opinions are precursors to many business decisions. Similar to the stop-words, sentiment analysis domain depends on list of words that describe the affect of the writer. (Nielson, 2011) explains how ANEW list is used to classify the opinions of users as negative, neutral or positive. (Wang et. al., 2011) applied the sentiment analysis concepts to topics rather than the actual opinions and gleaned how the discussion would follow the sentiment of a topic.

(Pang, 2008) also looked at the user reviews and the possible mistakes that can occur during the data entry when it comes to providing ratings. (Goldberg et. al, 2007, Hopkins et. al, 2007) looked at the opinions of democrat voters and how they felt about the presidential elections. (Bansal et. al., 2008) looked at the long-term aspect of sentiment analysis in the political domain where the voters can get a long-term view of how the politicians act during their tenure. Other projects have as a long-term goal the clarification of politicians' positions.
(Jin et. al., 2007) used applied sentiment analysis concepts to detect inappropriate ads. (Cheong et. al., 2011) tackled the problem of finding content related to terrorism. The work looked at the sentiments of civilians and how the twitter data can be harvested to glean such information. More work has been done in the field of twitter harvesting by (Saif et. al., 2012, Agrawal et. al., 2011 and Rosenthal et. al., 2017). Twitter data was also used to build a corpus that could be specific to twitter as explained by (Pak et. al., 2010). Finally, (Cobb et. al., 2013) has applied sentiment analysis to smoking-cessation techniques given certain drug choices.

## 3.2. NLP Application in Medical Sciences

Work done by (Youyou et. al., 2015) focuses on the Facebook platform where the authors compare the human perception to that being gleaned from the social media. Specifically, the authors judge the judgements of users by the numbers of "likes" the users press. On the other hand, they build two samples of more than 14 thousand users where they ask the users' friends to rate the judgement of the user. The results showed that the results gleaned from Facebook likes were more accurate. The results once again provide the researches evidence that information gleaned from the social media can prove to be both valuable and accurate. The results do not reflect the comments that the users make on various sites/items which can grant more accuracy to the results. Moreover, there is no corpus involved in the process. Work done by (Rajput et. al., 2018) can be leveraged in this regards.

Work done by (Chen et. al., 2016) provides an overview of Big Data application to Psychology. The authors focus on the four steps necessary for such endeavours namely planning, acquisition, planning and analytics. They also provide three tutorials for the users. They introduce the user to the MapReduce framework that has garnered lot of attention lately. The paper also explains the concept of supervised and unsupervised learning as explained above. The work while providing an excellent overview does skip certain details that the user might need. Specifically, they glance over the pre-processing part of data processing and the many underlying details such as text normalization.

The work done by (Hann et. al., 2013) focuses on text normalization for the Out of Vocabulary words. Specifically, the users target words such as "smokin" and find a mechanism to convert it into "smoking". The work focuses primarily on SMS messages and also delves into a decent sample size for Twitter. The authors proposed method produces very encouraging results comparing it to a corpus obtained from New York Times. The OOV words are first normalized using a dictionary-based matching and based on the results the authors move further to test in a context setting. (Rajput et. al., 2018) shows results in a psychology/psychiatry context – specifically depression. One of the issues the authors do not look into is converting OOV synonyms into actual dictionary based words – as an example "imo" into "in my humble opinion". A similar approach of normalization was also taken by (Gordon et. al., 2009) where the authors construct a normalization dictionary for Weblogs. The work provides an excellent way of pre-processing blogs – another social media platform. The work is not domain specific and hence the ideas can be applied to the more domain specific context.

Work done at Stanford University (Kosinski et. al., 2016) looks at the concept of digital footprint of a user and two mathematical ways to analyse data. The work is based on R language (as opposed to Python) and helps predict real life outcomes. The work is based in the unsupervised learning domain and uses Facebook data as a case study. While the work is not focused in the medical/social sciences domain, it is an excellent introduction to digital footprint of a user and can come in very handy in detection symptoms specific to various conditions.

De Choudhry et. al. has done some work in detecting mental illness. The authors started the work by focusing on detecting post-partum depression. The authors chose reddit as their platform and studied the linguistic changes that happened in new mothers. They showed the prevalence of negative affect in certain cases among other results. The work was an excellent first step in this realm and was followed by (De Choudhry et. al, 2013, De Choudhry, Counts et. al, 2013, and De Choudhry, 2013) where they predict depression in Twitter users. One of the prerequisites of their work is that the users identify themselves as depressed and gave consent to follow their Twitter account. The next step in this field should be to detect symptom of depression from random set of tweets. Moreover, the DSM V guidelines should also be followed and brought in line with the social media text.

(De Choudhry et. al, 2016) also did some work in focusing on the anonymity factor of the social media and the disinhibitions that accompany such anonymity. Specifically, they looked at the 'throwaway' accounts that the users used to express their opinions. The authors also applied more Big Data and Machine Learning techniques to the Nutrition area where they used social media to understand dietary choices among the social media users (Pavalanathan et. al, 2015). While the work is not directly related to mental illness, it can be leveraged in various situations such as cases of eating disorders occurring with other conditions such as depression or bipolar.

(Saha et. al., 2017) modelled stress among a group of college students after cases of shooting on campus. The authors used reddit campus community and looked at the linguistic style of the students making posts to detect high level of stress after such traumatic incidents. They looked at 12 incidents of gun violence over a period of five years and analysed both the time and linguistic dimension of the posts. The work done gives a great platform to spring from and see whether clinical inferences can be done given such data.

It is worth mentioning the work in (O'Callaghan , 2012) where the authors choose YouTube platform and focus on detecting spam in the comments section. The authors use graph and network theory to look at bot (automatic programs) behaviour. Such work can be combined with other work and help researchers classify various videos from myriad of dimensions. Once such classification is accomplished, the researchers can look at the embedded comments and tie the content to user mental state and how they are expressing it.

While the aforementioned work analysed from a psychology point of view (also applicable to clinical part of psychiatry), (Montieth , 2016) present a list of projects that are underway in the realm of psychiatry. The authors look at the Big Data field from a medical sciences perspective and emphasize how Big Data in a clinical setting can provide benefits such as pinpointing rare events. The work done in this area ranges from the use of psychotropic drugs to comparing the risk of dementia in a certain age group.

# 4. Conclusion

Sentiment Analysis and Natural Language Processing offer a great opportunity to mental health practitioners to be able to mine text data from the web and detect symptoms that could be harbinger to various mental health issues. In this chapter, we have presented a detailed overview of NLP and Sentiment Analysis. In addition, we have given a brief synopsis of strength of Python language and the NLTK toolkit. Lastly, we have presented various applications in the sentiment analysis domain and application of NLP to the medical field. In the future, the author would like to apply sentiment analysis in detection of mental health diseases such as cyberbullying and depression.

### 4.1. **Future Research Directions**

To begin with, social media offers the best source of data for mental health practitioners. The data is produced by real users and the anonymity provided by the Internet allows the practitioners to gather authentic data that can provide valuable insights into the mindset of patients suffering from mental illness.

The social media allows patients to post comments using their language of choice. Such data should be segregated and compared to see whether there are any commonalities between patients irrespective of their cultural background. For example, an interesting question to ask will be: Are the underlying drivers of depression as reported by users suffering from depression in the US and an Arabic speaking country the same?

Furthermore, while English remains the language of choice for majority of the users on the Internet, there are subtle differences in the way people write their comments. Thus, an interesting area to look into is ways to segregate English speaking patients based on their cultural and geographical backgrounds and glean the differences. For example, do users hailing from India and Pakistan display the same symptoms as the users from United Kingdom?

Lastly, another area to look into will be the ability to glean the Socioeconomic Status (SES) of the users that are posting to the social media. The SES status is a valuable piece of information that the healthcare practitioners use to predict the onset of various diseases. One facet of SES is the level of education of the user who is posting.

### 4.2. Teaching Assignments

- Perform a literature review of the work done on cyber bullying and identify how sentiment analysis and NLP techniques can be helpful
- Perform a literature survey on various techniques that can rank the writing style of various users
- Research the bi-gram and tri-gram data made available by Google and how it can be helpful in application to the domain of healthcare analytics